\documentclass[letterpaper]{article} 
\usepackage{aaai23}  
\usepackage{times}  
\usepackage{helvet}  
\usepackage{courier}  
\usepackage[hyphens]{url}  
\usepackage{graphicx} 
\usepackage{natbib}  
\usepackage{caption} 
\usepackage{algorithm}
\usepackage{algorithmic}

\usepackage{soul}
\usepackage[utf8]{inputenc}
\usepackage{amsmath}
\usepackage{amsthm}
\usepackage{amssymb}
\usepackage{booktabs}
\usepackage{color}
\usepackage{multirow}
\usepackage{newfloat}
\usepackage{listings}
\DeclareCaptionStyle{ruled}{labelfont=normalfont,labelsep=colon,strut=off} 
\floatstyle{ruled}
\newfloat{listing}{tb}{lst}{}
\floatname{listing}{Listing}
\title{Adaptive Texture Filtering for Single-Domain Generalized Segmentation}
\author{
    Xinhui Li\textsuperscript{\rm 1}, Mingjia Li\textsuperscript{\rm 1},
    Yaxing Wang\textsuperscript{\rm 2}, Chuan-Xian Ren\textsuperscript{\rm 3}, Xiaojie Guo\textsuperscript{\rm 1}\thanks{Corresponding Author}
}
\affiliations{
    \textsuperscript{\rm 1}Tianjin University, Tianjin, China \\
    \textsuperscript{\rm 2}Nankai University, Tianjin, China\\
    \textsuperscript{\rm 3}Sun Yat-sen University, Guangdong, China\\

    \{lixinhui, mingjiali\}@tju.edu.cn, yaxing@nankai.edu.cn, rchuanx@mail.sysu.edu.cn, xj.max.guo@gmail.com
}

\usepackage{bibentry}

\begin{document}

\maketitle

\begin{abstract}
Domain generalization in semantic segmentation aims to alleviate the performance degradation on unseen domains through learning domain-invariant features. 
Existing methods diversify images in the source domain by adding complex or even abnormal textures to reduce the sensitivity to domain-specific features. However, these approaches depend heavily on the richness of the texture bank, and training them can be time-consuming.
In contrast to importing textures arbitrarily or augmenting styles randomly, we focus on the single source domain itself to achieve generalization.
In this paper, we present a novel adaptive texture filtering mechanism to suppress the influence of texture without using augmentation, thus eliminating the interference of \textit{domain-specific} features. 
Further, we design a hierarchical guidance generalization network equipped with structure-guided enhancement modules, which purpose is to learn the \textit{domain-invariant} generalized knowledge. Extensive experiments together with ablation studies on widely-used datasets are conducted to verify the effectiveness of the proposed model, and reveal its superiority over other state-of-the-art alternatives.
\end{abstract}

\section{Introduction}
The powerful data representation capability endows deep neural networks with remarkable achievements in a series of computer vision tasks, \emph{e.g.}, semantic segmentation, object detection, and scene understanding. These successes are usually based on the assumption that models should be trained and tested on data having the same underlying distribution. However, in realistic scenarios, it is impractical to collect sufficient data that covers the entire space. Even more, some deployed domains are completely inaccessible. For instance, collecting images under all possible conditions and scenarios are impossible in the self-driving system. Therefore, as a representative line, domain generalization (DG) \cite{Carlucci2019DomainGB,Qiao2020LearningTL,Peng2021GlobalAL} is extensively investigated to address a fundamental flaw: the network exhibits obvious performance degradation in the face of out-of-distribution or unseen-domain data. It is also known as the \emph{distribution shift} \cite{MorenoTorres2012AUV} issue. 

\begin{figure}[t]
\centering
\includegraphics[width=1.0\columnwidth]{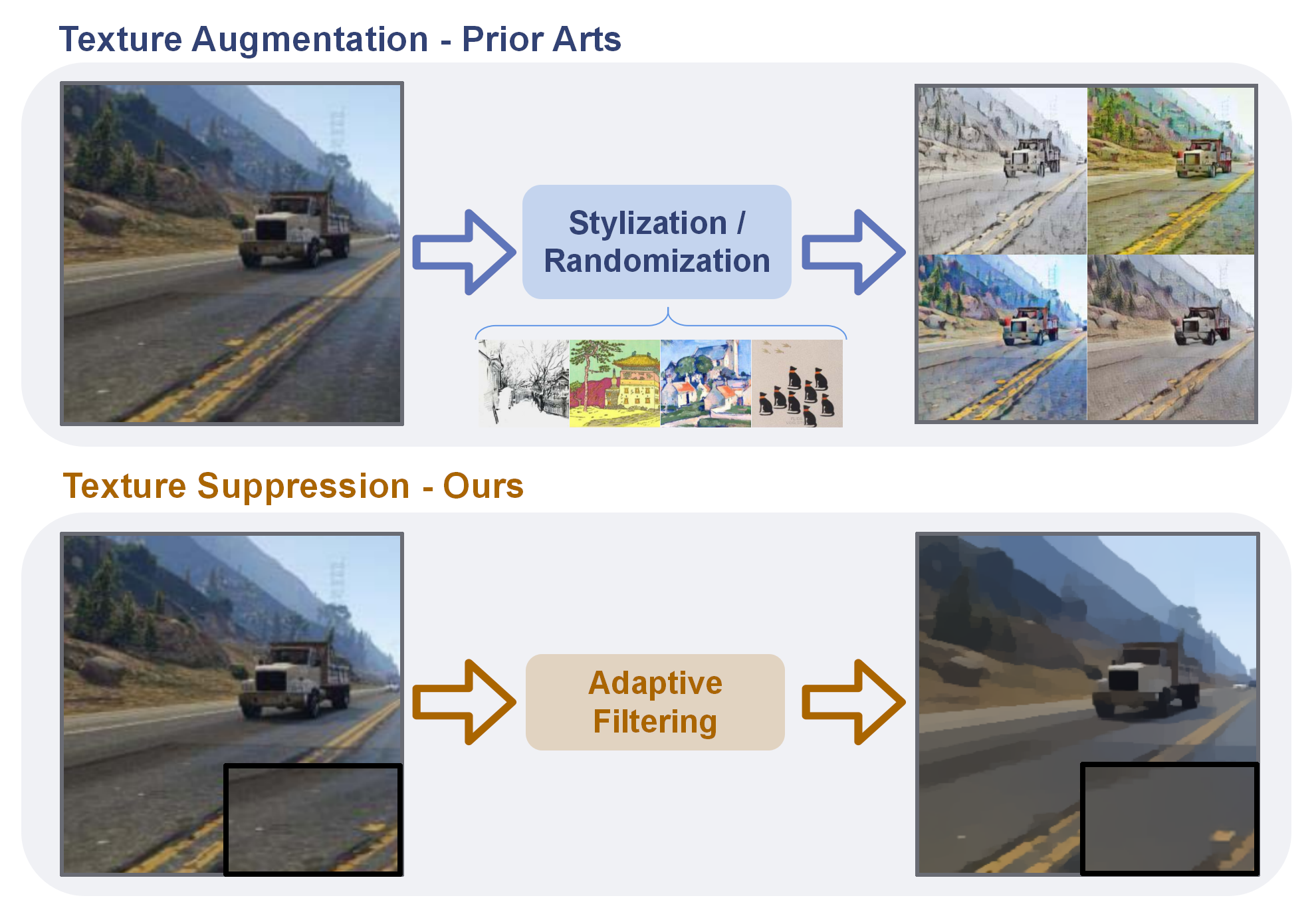} 
\caption{Illustration of existing DG methods \cite{Peng2021GlobalAL} using texture augmentation, and our proposed approach using texture suppression.
}
\label{intro}
\vspace{-5pt}
\end{figure}

Since \emph{data in the target domain is inaccessible during the training process}, current DG methods advocate alleviating the distribution shift by reducing the domain gap (\emph{e.g.}, appearance, style, or texture).
These approaches can be divided into two main categories, \emph{i.e.}, multi-source DG and single-source DG. Multi-source DG schemes \cite{Li2018DomainGW,Peng2019MomentMF,Matsuura2020DomainGU,Zhu2021SelfsupervisedUD} attempt to align multiple available source domains through minimizing the divergence among domains. 
However, the alignment is typically based on the accessibility of multiple domain distributions that are not always available. 
To mitigate the pressure from data, another branch of DG approaches \cite{Tobin2017DomainRF,Volpi2018GeneralizingTU,Qiao2020LearningTL,Peng2021GlobalAL,Wang2021LearningTD} using single-source has been studied to achieve the goal. Some of recent methods \cite{Ulyanov2017ImprovedTN,Pan2018TwoAO} investigate the normalization to standardize feature distributions, while the strategies \cite{Prakash2019StructuredDR,Kang2022CVPR,Tjio2022AdversarialSH} focus on the style transformation to diversify image styles. The other works \cite{Tobin2017DomainRF,Ulyanov2017ImprovedTN,Yue2019DomainRA,Peng2021GlobalAL} execute the domain randomization for texture synthesis. As shown in Fig.~\ref{intro}(top), some works mainly focus on diversifying the appearance (\emph{e.g., texture}) \cite{Peng2021GlobalAL}.
However, these data augmentation methods are limited by the authenticity and randomness of added textures, and can hardly cover all the appearances that may arise in the target domain. To handle this problem, we investigate the necessity of importing additional textures (or appearances) through the complex transformation. Namely, can we find a straightforward way to learn domain-invariant features by explicitly filtering textures from images? 

In this paper, we concentrate on the task of single-domain generalization and introduce a method from a novel perspective.
We work on \emph{texture suppression}, as shown in Fig.~\ref{intro} (bottom), which is more flexible and simpler than the data augmentation operation.
We propose an adaptive filtering mechanism (AFM) to screen out the texture component from images, aiming to explicitly mitigate the effect of texture.
More concretely, our AFM can break through the deficiency of traditional filtering which requires setting the filtering level manually. 
The proposed AFM can adaptively generate texture filtering intensity parameters by predicting the style of the input image, and then producing a content-dependent image. 
Moreover, to further explore the domain-invariant feature representation in the content-dependent image, we design a hierarchical guidance generalization network (HGGN). HGGN is customized with structure-guided enhancement modules, which purpose is to extract domain-invariant features and enhance spatial information learning ability.
The major contributions of this work are summarized as follows:
\begin{itemize}
    \item We propose a novel adaptive filtering mechanism to generate the content-dependent image, aiming to eliminate the interference of domain-specific features. 
    \item We design a hierarchical guidance generalization network, which can implicitly guide the network to learn domain-invariant features.
    \item Extensive experiments are conducted on various datasets to validate the effectiveness of our method and show the superiority of our approach over other state-of-the-art.
\end{itemize}

\section{Related Work}

\subsection{Unsupervised Domain Adaptation}
Manually labeling semantic segmentation data is extremely time-consuming and laborious. Thus unsupervised domain adaptation (UDA) \cite{luo2022towards} has drawn growing attention, which can liberate the requirement of annotation. Basically, the manner of UDA is by narrowing the distribution gap between the automatically labeled synthetic data and unlabeled real data to achieve reasonable performance in the real domain. For instance, several works utilize the maximum mean discrepancy (MMD) \cite{Geng2011DAMLDA,Long2015LearningTF} to minimize a certain distance measure between domains and then align the distributions of different datasets. However, because of the insufficient capacity of MMD minimization, domain alignment is not always guaranteed. 
Other popular approaches leveraging the idea of adversarial learning \cite{Luo2019TakingAC,Tsai2019DomainAF,Zhang2020JointAL,Shermin2021AdversarialNW} are applied by adopting domain discriminators. 
Meanwhile, style transfer and image-to-image translation \cite{Zhu2017UnpairedIT,huang2017arbitrary,Lee2021UnsupervisedDA} methods are also studied to the UDA with the goal of converting the style of the source data and possibly getting closer to the target data.
Unfortunately, the distribution discrepancy cannot be estimated from the source domain alone when real data is not accessible during training. In other words, hardly these techniques can cope with domain generalization.

\begin{figure*}[t]
\centering
\includegraphics[width=1.0\textwidth]{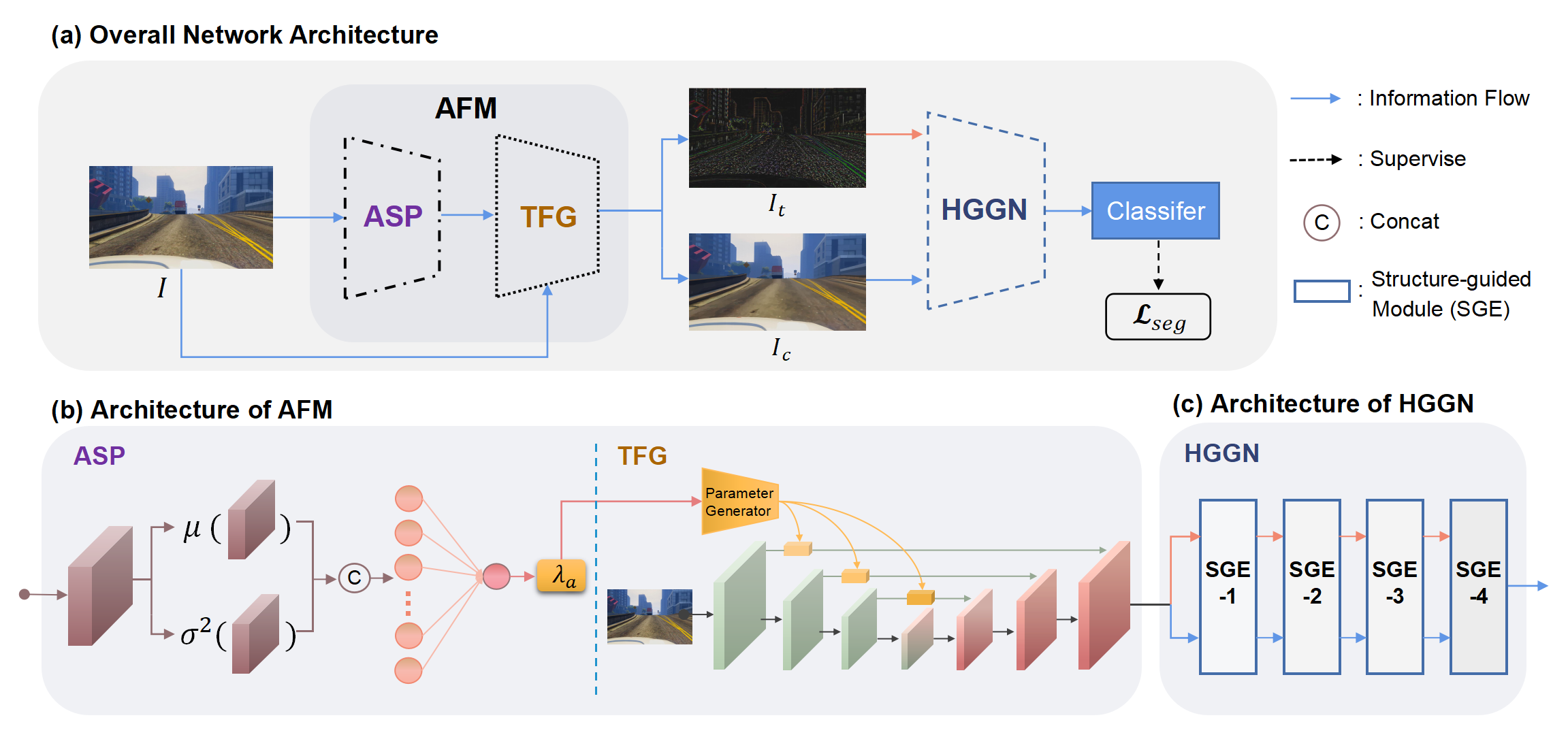} 
\caption{(a) Network architecture of the proposed method, including the adaptive filtering mechanism (AFM) and hierarchical guidance generalization network (HGGN). The AFM consists of the adaptive strength predictor (ASP) module and the texture filtering generator (TFG) module. (b) ASP predicts the filtering intensity parameter $\lambda_{a}$ by calculating the mean $\mu$ and variance $\sigma^2$ of the extracted features. TFG is responsible for generating the content-dependent image $I_{c}$ and texture-dependent image $I_{t}$. (c) HGGN composed of structure-guided enhancement modules (SGE) guides the network to learn generalized representations.}
\label{network}
\end{figure*}

\subsection{Domain Generalization}
Unlike UDA, domain generalization (DG) methods cannot access the target domain. The goal of DG is to train only on the source domain and generalize on target domains. The paradigm of DG methods is basically to learn generalized feature representations.
Existing DG methods can be roughly divided into two categories: 1) Multiple-source DG \cite{Matsuura2020DomainGU,Wang2021GeneralizingTU,Liu2021FedDGFD,Li2021SemanticSW}, and 2) Single-source DG \cite{Volpi2018GeneralizingTU,Qiao2020LearningTL,Peng2021GlobalAL,Wang2021LearningTD,Xu2022DIRLDR}. 

\textbf{Multiple-source DG} methods attempt to explore common latent representations among the available source domains to improve the generalization capability.
Inspired by the similar idea of feature alignment on UDA, many approaches mainly utilize explicit feature distribution constraints \cite{Peng2019MomentMF,Zhu2021SelfsupervisedUD} or feature normalization \cite{Nam2018BatchInstanceNF,Jia2019FrustratinglyEP} to learn domain-invariant feature representations on multiple source domains.
However, due to the uncertainty of target domains, the feature alignment method has the risk of over-fitting source domains and leads to poor generalization ability of the unseen domain.
Moreover, various methods such as adversarial feature learning \cite{Li2018DomainGW}, meta-learning \cite{Li2019EpisodicTF,Shu2021OpenDG,Kim2022PinTM}, and metric learning \cite{Motiian2017UnifiedDS,dou2019domain} are also proposed to boost the generalization performance. 
In this paper, we focus on single-source DG.

\textbf{Single-source DG} methods strive to solve the generalization issue from the perspective of augmenting source data, such as domain randomization \cite{Yue2019DomainRA} or style transformation \cite{Prakash2019StructuredDR,Peng2021GlobalAL} to diversify the distribution of single source domain. However, it is inflexible to perform complex adversarial generation or unstable stylized transformation on the source image. 
Another studies focus on exploiting latent feature representation through feature normalization, such as instance normalization, batch normalization, and many other extended normalization methods \cite{Pan2018TwoAO,choi2021robustnet,peng2022semantic}. These works show that normalization is effective to solve the DG problem in preventing the over-fitting of training data.
Although numerous studies have been proposed to solve the generalization problem, most of the DG methods mentioned above mainly focus on image classification \cite{Kang2022CVPR}, and a few researches \cite{Yue2019DomainRA,choi2021robustnet} on semantic segmentation.
However, for the field of autonomous driving, semantic segmentation in DG is valuable and practical in the real world.
Our work belongs to the single-source DG and focuses on addressing the practical application of DG in semantic segmentation.

\section{Methodology}
This work focuses on how to solve the single-domain generalization problem: a model is trained on the source domain, expecting to generalize well on many unseen real-world domains. 
Due to the effect of domain-specific textures in generalization, as previously analyzed, our method aims to alleviate the influence of domain-specific component through texture filtering. 
For this purpose, we propose a novel adaptive filtering mechanism to obtain the content-dependent component. 
This is a straightforward and efficient way for the network to learn feature representation without the interference of texture.
In addition, we design a hierarchical guidance generalization network to implicitly guide the network to learn domain-invariant features.

\subsection{Overall Architecture}
The overall architecture of our method is depicted in Fig.~\ref{network}.  
Given an input image $I$, we first filter its texture by the adaptive filtering mechanism (AFM). Within it, the adaptive strength predictor (ASP) estimates a filtering intensity parameter $\lambda_{a}$ determining the strength of the texture filtering generator (TFG). The TFG produces the texture-dependent image $I_{t}$ and content-dependent image $I_{c}$. The hierarchical guidance generalization network (HGGN) stacked by structure-guided enhancement (SGE) modules is linked after the AFM to learn domain-invariant feature representations under contour supervision.

\begin{figure*}[t]
\centering
\includegraphics[width=1.0\textwidth]{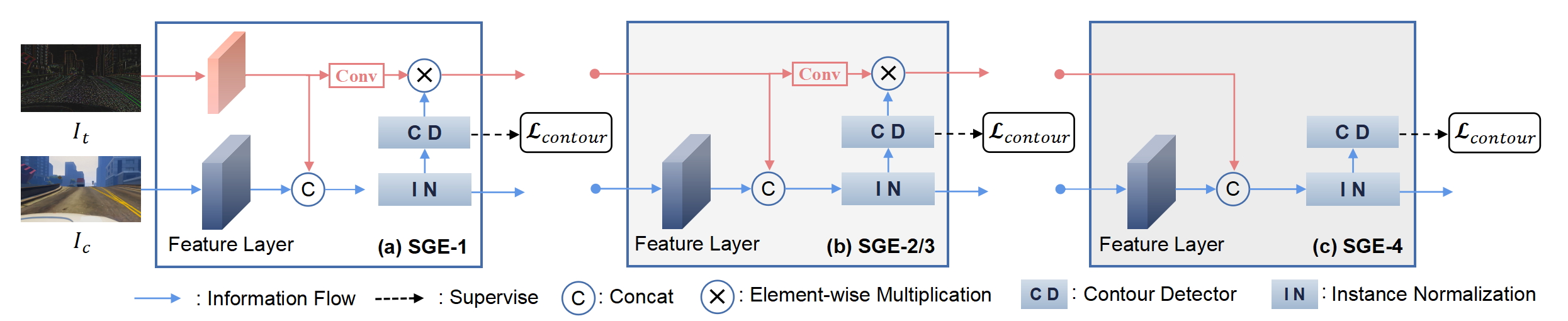} 
\caption{Network architecture of Structure-guided Enhancement Modules (SGE), which constitute the HGGN.}
\vspace{-5pt}
\label{HGGN}

\end{figure*}

\subsection{Adaptive Filtering Mechanism}
The key to ensuring the generalization performance is the ability to generate the content-dependent component that allows the network to learn domain-invariant features directly.
To this end, the adaptive filtering mechanism is proposed to separate the image into content-dependent component $I_{c}$ and texture-dependent component $I_{t}$, which contains the adaptive strength predictor (ASP) and texture filtering generator (TFG).

\subsubsection{\textit{Adaptive Strength Predictor (ASP)}}
ASP aims to generate the filter strength value based on the style of the input image.
Specifically, we implement a feature extraction module containing three convolution layers and then calculate the mean $\mu$ and variance $\sigma^2$ of the extracted features across the spatial dimension of each channel, respectively. Due to the effectiveness of the statistics in representing image information \cite{huang2017arbitrary,zhou2021domain}, the value of mean and variance are concatenated to represent the style of the input image.
Then, the value of filter strength $\lambda_{a}$ can be obtained through a fully connected layer, as shown in Fig.~\ref{network}(b).
For more detail, in the training period, we add the $\lambda_{a}$ with an error term $\epsilon$, whose value is obtained through the sampling of the normal distribution with a mean of 0 and a variance of 1, and truncated at 1.5. This operation can effectively improve the robustness of the network during training and is removed in the testing.
Finally, after the processing of ASP, the filter strength value $\lambda_{a}$ required by TFG can be automatically generated based on the style of the input image.

\subsubsection{\textit{Texture Filtering Generator (TFG)}} 
TFG is designed to filter the texture component from the image while keeping the structure. 
Motivated by the effectiveness of image filtering or smoothing \cite{Guo2020MutuallyGI,Li2022DeepFS} on texture removal, we adopt smoothing operation as texture filter.
As shown in Fig.~\ref{network}(b), we utilize a U-shape network as the backbone of our smoothing generator. As noticed by previous art \cite{filterbank}, the smoothing strength can be manipulated by inserting different convolution modules in the skip-connection. 
According to the smoothing strength $\lambda_{m}$ given we first generate convolution kernels, and then use the generated convolution kernel to perform structure-preserving smoothing.
Note that our TFG is pre-trained with labels from the training split of the GTA5 dataset, and then the parameters of TFG is fixed when training the segmentation network.

We extract semantic boundaries from segmentation annotations as training guidance to protect semantically meaningful edges, which is significant for segmentation. Given input image $I$ and semantic boundary $G$, the loss function of TFG can be expressed as follows:
\begin{equation}
    \label{TFG_loss}
    \mathcal{L}_{smooth} = \| I - S \|^{2}_{2} + \lambda_{m} 
    \left\|\frac{\nabla S}{\nabla I + G + \epsilon} \right\|^{2}_{2},
\end{equation}
where $S$ is the smoothed output, $\| \cdot \|_2$ represents the $\ell_2$ norm, $\nabla \cdot$ represents the gradient of an image, $\epsilon$ is a small constant to avoid zeros in the denominator (set to 0.005 empirically) and $\lambda_{m}$ is the smoothing strength. 
When training the TFG, we randomly sample $\lambda_{m} \in [0, 4] $ from a uniform distribution as the smoothing strength. The first term in $\mathcal{L}_{smooth}$ regularizes the outputs to be consistent with the input image, while the second term tends to penalize the texture. That is to say, the higher $\lambda_{m}$ is, the more smooth the output will be.  
Finally, through the cooperation of ASP and TFG, 
we regard the the smoothed image $S$ without texture as the content-dependent image $I_{c}$.
Meanwhile, the texture-dependent image $I_{t}$ is expressed as $I_{t}=I-I_{c}$. In Fig.~\ref{complementary}, we show some samples of content-dependent images and texture-dependent images.

\subsection{Hierarchical Guidance Generalization Network}
\begin{figure}[t]
\centering
\includegraphics[width=1.0\columnwidth]{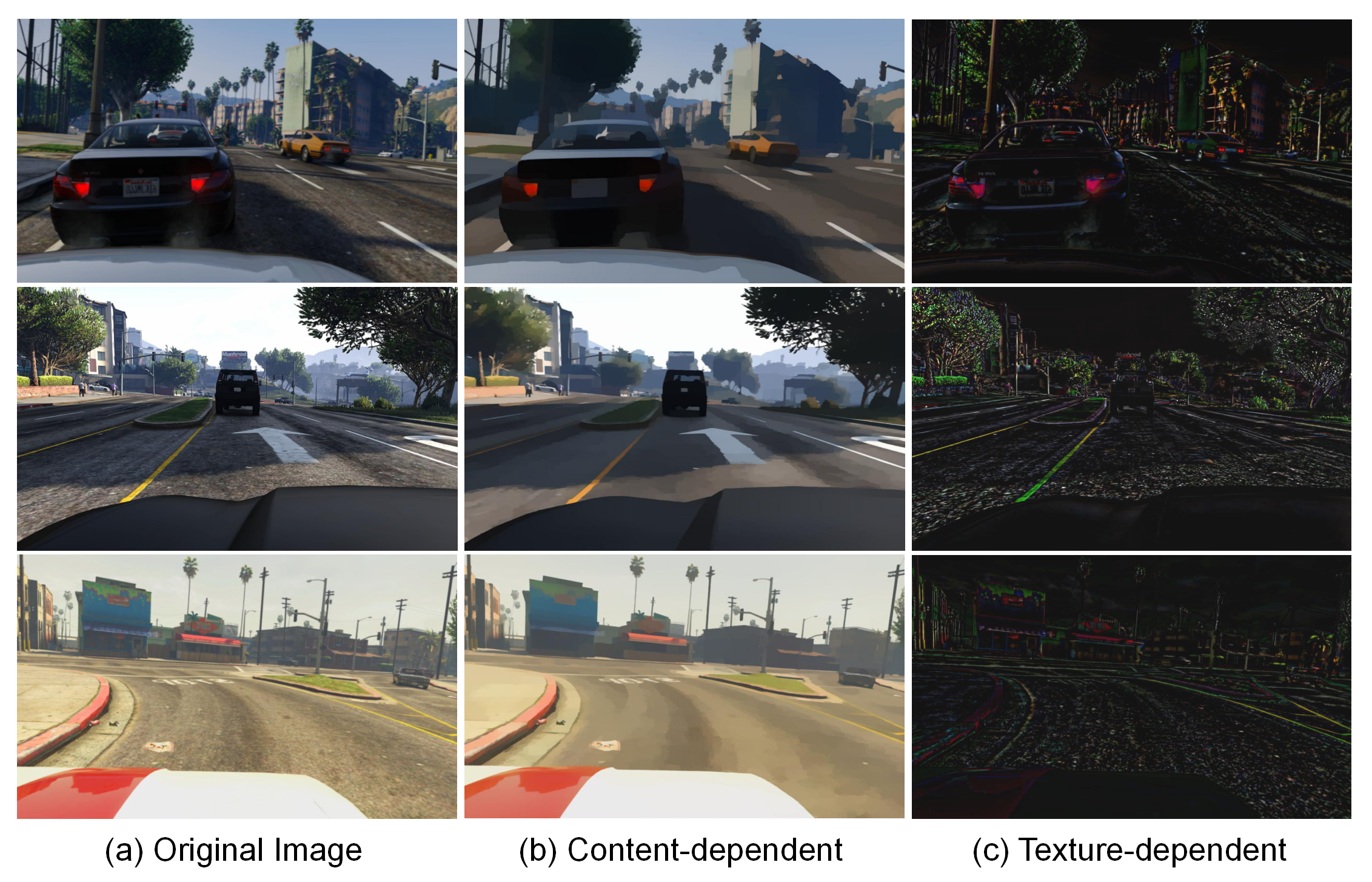} 
\caption{Samples of content-dependent images $I_{c}$ and texture-dependent images $I_{t}$ generated from the AFM in the training period. Original images are from the GTA5 dataset.}
\label{complementary}
\vspace{-5pt}
\end{figure}

To learn the generalized representation of the content-dependent image $I_{c}$ effectively, we propose a hierarchical guidance generalization network (HGGN) including structure-guided enhancement modules (SGE), as shown in Fig.~\ref{HGGN}.  
Firstly, we utilize the $I_{c}$ generated from the AFM as the input of the HGGN. Although the filtering operation can preserve the contour of objects, some internal structure information of the object would be removed during this processing. However, the internal structure information is also helpful for the segmentation task~\cite{ronneberger2015u}. To this end, we also extract the internal structure from the texture-dependent image $I_{t}$. In Fig.~\ref{complementary}, it can be clearly observed that $I_{t}$ contains some internal structure information of the object.
Therefore, we design SGE modules to fuse the extracted feature from $I_{c}$ and $I_{t}$, which can better guide the network to learn the spatial feature representation.  

More specifically, there are four SGE modules in the HGGN, as shown in Fig.~\ref{HGGN}. To ensure that internal structure information in $I_{t}$ is better preserved, we adopt concatenate to fuse the features. 
Subsequently, we normalize the fused features through instance normalization (IN) to extract the style-normalized feature representations.
Many previous works like \cite{Ulyanov2017ImprovedTN,Jin2020StyleNA} have proved the effectiveness of IN on style normalization, but the spatial feature representation might be weakened during IN.
However, in the DG semantic segmentation task, the spatial features after IN need to be highlighted due to the similar structural information among different domains.
Specifically, for this purpose, we design a contour detector (CD). CD generates the predicted contour map $y$ through three convolution layers and then calculates the contour loss $\mathcal{L}_{contour}$ with the contour ground truth, which is obtained from the semantic label. 
Meanwhile, the contour map $y$ is also utilized to enhance the spatial feature representation of the extracted feature from $I_{t}$ by element-wise multiplication.
A convolution layer is used to change the channel dimension, namely the ``Conv" in Fig.~\ref{HGGN}.
The supervised training of contour detection can explicitly assist the model in learning the domain-invariant (shape and spatial) information. We use the class-balanced cross-entropy loss as the contour loss to supervise the predicted contour map.
For a predicted contour map $y$, the contour loss $\ell_{\text {contour}}$ can be written as:
\begin{eqnarray}
 \begin{aligned}
\ell_{\text {contour}}=&-\beta \sum_{j \in Y_{+}} \log \sigma\left(y_{j}=1\right)\\
&-(1-\beta) \sum_{j \in Y_{-}} \log \sigma\left(y_{j}=0 \right),
 \end{aligned}
\end{eqnarray}
where $\sigma(\cdot)$ is the sigmoid function. $\beta=\left|Y_{-}\right| /\left(\left|Y_{-}\right|+\left|Y_{+}\right|\right)$, where $\left|Y_{-}\right|$ and $\left|Y_{+}\right|$
denote the contour and non-contour in the ground truth $Y$. The total contour loss $\mathcal{L}_{\text {contour}}$ is obtained by the sum of $\ell_{\text {contour}}$ from four SGE modules.

In addition to the SGE, the classifier is responsible for generating semantic segmentation results. The segmentation loss $\mathcal{L}_{\text {seg}}$ given by standard cross-entropy loss is defined as:
\begin{eqnarray}
\begin{gathered}
\mathcal{L}_{\text {seg }}=-\sum_{h, w} \sum_{c=1}^{C} y_{s}^{(h, w, c)} \log p_{s}^{(h, w, c)},
\end{gathered}
\end{eqnarray}
where $p_{s}^{(h, w, c)}$ denotes the predicted semantic segmentation result. $y_{s}^{(h, w, c)}$ is the semantic ground truth label and $C$ is the number of classes. Combining the above contour loss term $\mathcal{L}_{\text {contour}}$ yields our final objective function:
\begin{eqnarray}
\begin{gathered}
\mathcal{L}_{\text {total }}=\mathcal{L}_{\text {seg}}+\alpha \cdot \mathcal{L}_{\text {contour}},
\end{gathered}
\end{eqnarray}
where we use hyper-parameter $\alpha$ to balance the importance of semantic segmentation loss and contour loss. Here, $\alpha$ is empirically set to 2.5.

\section{Experiments}
In this section, we first describe the experimental setup in detail and then reveal the advance of our design in comparison with other state-of-the-art on different real-world datasets. Finally, the effectiveness of our designed modules is analyzed in ablation studies.

\subsection{Dataset Description}
To verify the generalization ability of our method, we conduct experiments on extensive datasets following the common protocol adopted by prior works. There are two synthetic datasets (\emph{e.g.}, GTA5 and SYNTHIA) and three real-world datasets (\emph{e.g.}, Cityscapes, BDD-100K, and Mapillary) in the experiments. 

\subsubsection{Source Domain Datasets} 
For source domain datasets, GTA5 \cite{Richter2016PlayingFD} and SYNTHIA \cite{Ros2016TheSD} are used with automatically generated annotations during training, respectively.
GTA5 is a large synthetic dataset containing 24966 urban scene images of size 1914$\times$1052. It is rendered by Grand Theft Auto V game engine and automatically annotated into different semantic categories by pixel.
SYNTHIA is a synthetic dataset with pixel-level semantic annotation. We use the SYNTHIA-RAND-CITYSCAPES subset in our experiments, which contains 9,400 synthetic images with a high resolution of 1280$\times$760. 

\subsubsection{Target Domain Datasets}
To evaluate the generalization capability, three real-world datasets: Cityscapes \cite{Cordts2016TheCD}, BDD-100k \cite{Yu2020BDD100KAD}, and Mapillary \cite{Neuhold2017TheMV} are adopted during testing. We only utilize the validation set of these datasets to test the performance of our model for comparison with other approaches.
Cityscapes is a large-scale semantic segmentation dataset collected from 50 different cities in street scenarios. It contains a training set with 2,975 images and a validation set with 500 images.
BDD-100k is an urban driving scene dataset collected from various locations in the US, which has 7000 training images and 1000 validation images.
Mapillary is a diverse street view dataset with annotations of 66 classes, and we only use the overlapped classes with the synthetic datasets. The training and validation sets contain 18,000 and 2,000 images, respectively.

\begin{figure*}[t]
\centering
\includegraphics[width=1.0\textwidth]{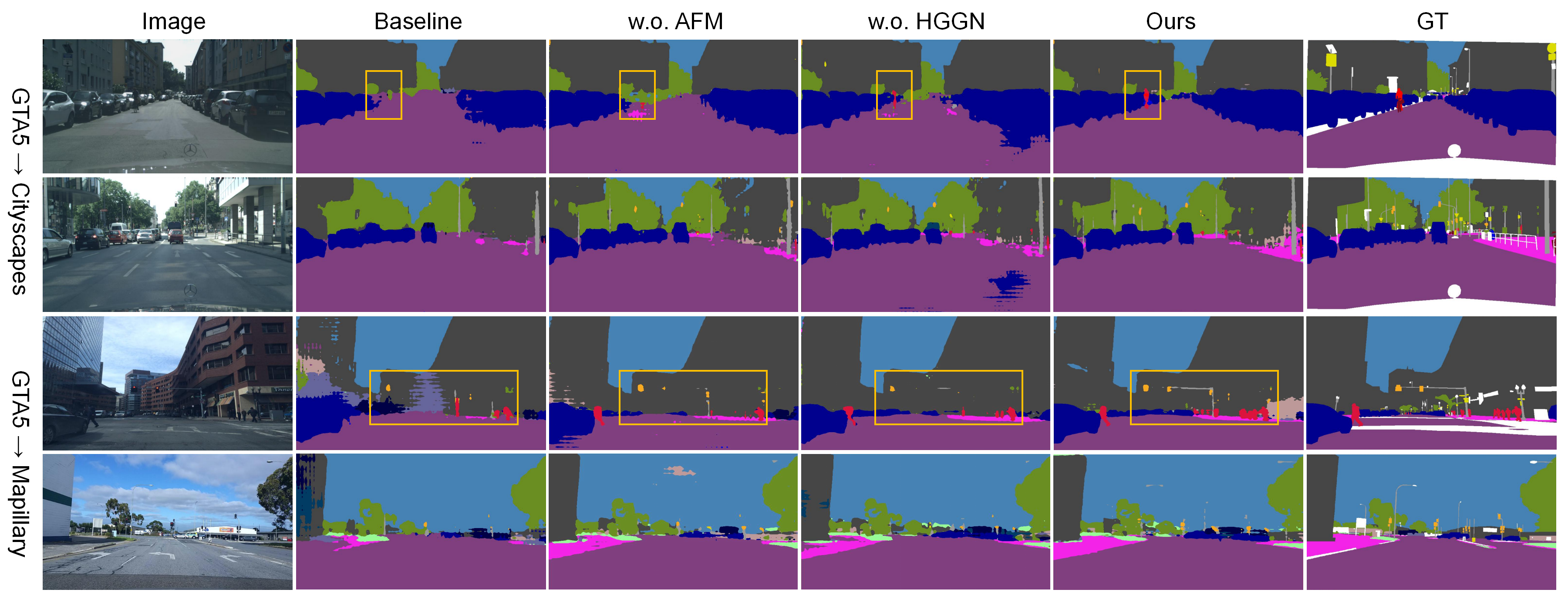} 
\caption{Qualitative domain generalization results on semantic segmentation. The model is trained on GTA5 \cite{Richter2016PlayingFD} and then generalized to Cityscapes \cite{Cordts2016TheCD} and Mapillary \cite{Neuhold2017TheMV} with ResNet-101.}
\vspace{-5pt}
\label{result}
\end{figure*}

\subsection{Implementation Details}
Following the current state-of-the-art, we train our model on the synthetic datasets and then test on the unseen real-world datasets. ResNet-50 and ResNet-101 \cite{He2016DeepRL} are used as the backbone, respectively. The baseline model is a segmentation network \cite{chen2018encoder} which follows the same architecture as prior works \cite{choi2021robustnet}. We use pre-trained parameters on ImageNet to initialize the model except for the classifier. In the training phase, we use Stochastic Gradient Descent (SGD) optimizer with a batch size of 2, a momentum of 0.9, and a weight decay of 0.0005.
The initial learning rate is set to 2.5e-4 and follows the poly learning rate policy with the power of 0.9. The training iteration is set to 200,000 for all involved models. We implement our method with PyTorch and use a single NVIDIA RTX3090 with 24 GB memory. The PASCAL VOC Intersection over Union (IoU) \cite{Everingham2014ThePV} is used as the evaluation metric to measure the segmentation performance, and mIoU is the mean value of IoUs across all categories.  During the training, we use random cropping and flipping. Meanwhile, we also adopt augmentations by randomly changing color, brightness, sharpness, and contrast.
\begin{table}[t]
\centering
\setlength{\tabcolsep}{2pt}
    \resizebox{\linewidth}{!}{
    \begin{tabular}{c|cccc}
    \toprule
    Methods (GTA5) &Mapillary  & BDD-100k & Cityscapes &Avg. \\
    \midrule
    \midrule
    \multicolumn{5}{c}{\emph{ResNet-101}}  \\
    \midrule
    Baseline &31.50 &27.85 &30.06 &29.80 \\
    \midrule
    IBN-Net (ECCV'18) &37.94 &38.10 & 37.23&37.76 \\
    \midrule
    DRPC (ICCV'19) &38.05 &38.72 & 42.53&39.77 \\
    \midrule
    ISW (CVPR'21)&39.05 &38.53 & 42.87&40.15\\
    \midrule
    GTR (TIP'21) &39.10 &39.60 &43.70 &40.80\\
    \midrule
    SAN-SAW (CVPR'22) &40.77 &41.18 &\textbf{45.33} &42.43\\
    \midrule 
    Ours &\textbf{45.59} &\textbf{42.31} &44.83 &\textbf{44.24} \\
    \midrule    
    \multicolumn{5}{c}{\emph{ResNet-50}}  \\
    \midrule
    Baseline &29.28 &25.31 &28.97&27.85 \\
    \midrule
    IBN-Net (ECCV'18) &37.75 &32.30 & 33.85&34.63 \\
    \midrule
    DRPC (ICCV'19) &34.12 &32.14 & 37.42&34.56\\
    \midrule
    ISW (CVPR'21) &40.33 &35.20 & 36.58&37.37 \\
    \midrule
    GTR (TIP'21) &34.52 &33.75 & 37.53&35.27 \\
    \midrule
    SAN-SAW (CVPR'22) &41.86  & 37.34 & 39.75&39.65\\
    \midrule
    Ours &\textbf{42.82} &\textbf{38.79} &\textbf{39.91} &\textbf{40.51}
 \\
    \bottomrule
    \end{tabular}
    }
    \caption{The comparison in mIoU ($\%$) with other DG methods of the ResNet-101 and ResNet-50. The model is trained on the GTA5 dataset and generalized to real-world datasets.}
    \label{tab:resnet101}
    \vspace{-10pt}
\end{table}

\begin{table}[t]
\centering
\setlength{\tabcolsep}{2pt}
    \resizebox{\linewidth}{!}{
    \begin{tabular}{c|cccc}
    \toprule
    Methods (SYN) &Mapillary  & BDD-100k & Cityscapes &Avg. \\
    \midrule
    \midrule
    \multicolumn{5}{c}{\emph{ResNet-101}}  \\
    \midrule
    Baseline &21.84 &25.01 & 23.85 &23.57 \\
    \midrule
    IBN-Net (ECCV'18) &36.19 & 36.63 &34.18 &35.67 \\
    \midrule
    DRPC (ICCV'19) &34.12 &34.34 &37.58 &35.35\\
    \midrule
    ISW (CVPR'21) &35.86 &33.98 &37.21 &35.68 \\
    \midrule
    GTR (TIP'21) &36.40 &35.30 &39.70 &37.13 \\
    \midrule
    SAN-SAW (CVPR'22) &37.26 &35.98 &40.87 &38.04\\
    \midrule
    Ours &\textbf{39.10} &\textbf{36.87} &\textbf{41.32} &\textbf{39.10}
 \\
 \midrule
 \multicolumn{5}{c}{\emph{ResNet-50}}  \\
     \midrule
    Baseline &21.79 &24.50 & 23.18 &23.16 \\
    \midrule
    IBN-Net (ECCV'18) &32.16 & 30.57 &32.04 &31.60 \\
    \midrule
    DRPC (ICCV'19) &32.74 &31.53 &35.65 &33.31\\
    \midrule
    ISW (CVPR'21) &30.84 &31.62 &35.83 &32.76 \\
    \midrule
    GTR (TIP'21) &32.89 &32.02 &36.84 &33.92 \\
    \midrule
    SAN-SAW (CVPR'22) &34.52  &35.24 &38.92&36.23\\
    \midrule
    Ours &\textbf{37.14} &\textbf{36.07} &\textbf{39.48} &\textbf{37.56}
   \\
    \bottomrule
    \end{tabular}
    }
    \caption{The comparison in mIoU ($\%$) with other DG methods of the ResNet-101 and ResNet-50 trained on the SYNTHIA (SYN) and generalized to real-world datasets.}
    \vspace{-10pt}
    \label{tab:synthia}
\end{table}

\subsection{Comparison with State-of-the-Art}
We compare our method with state-of-the-art in domain generalization for semantic segmentation. 
To evaluate the effectiveness of our method, extensive experiments are conducted to show the generalization capability in the unseen domains.
Table ~\ref{tab:resnet101} summarizes the comparison results with recent state-of-the-art on the semantic segmentation task including IBN-Net \cite{Pan2018TwoAO}, DRPC \cite{Yue2019DomainRA}, ISW \cite{choi2021robustnet}, GTR \cite{Peng2021GlobalAL}, and SAN-SAW \cite{peng2022semantic}.  
Benefiting from the texture suppression in AFM and domain-invariant representation learning in HGGN, our method outperforms most methods on real-world datasets.
Furthermore, we also adopt a more lightweight backbone network ResNet-50 to verify the generalization performance. 
It should be highlighted some other methods encounter lower improvement when switching the backbone network from ResNet-50 to ResNet-101, and even obtain degraded performance on the Mapillary dataset.
This indicates that the alternatives cannot overcome the generalization problem on the heavier model, while ours still maintains significant improvements. 
In addition, Table \ref{tab:synthia} shows the superior results trained on another synthetic dataset (SYNTHIA) and then tested on the same real-world datasets.
Extensive experiments demonstrate that our method can provide robust representations by using our proposed adaptive filtering mechanism and hierarchical guidance generalization network.

\subsection{Ablation Study}
We investigate the proposed modules including AFM and HGGN to find out how they contribute to the generalization ability of segmentation. Moreover, we study the sensitivity of the hyper-parameter $\alpha$ in the objective function. 

\begin{table}[t]
\centering
   \resizebox{\linewidth}{!}{
    \begin{tabular}{c|cccc}
    \toprule
    Settings & Mapillary & BDD-100k & Cityscapes &Avg. \\
    \midrule
    \midrule
    \textbf{w/o} AFM &39.72  & 35.49 &38.24 &37.82\\
    \midrule
    Fixed level ($\lambda_{a}$=1) &42.34 &41.02 &42.96 &42.11\\
    \midrule
    Fixed level( $\lambda_{a}$=2) &41.13 &39.28 &40.27 &40.23\\
    \bottomrule
    \end{tabular}
   }
    \caption{Ablation study of adaptive filtering mechanism (AFM). The model is trained on the GTA5 with ResNet-101 and validated on three real-world datasets.}
    \vspace{-5pt}
    \label{tab:ablationAFM}
    
\end{table}

\subsubsection{\textbf{Effect of Adaptive Filtering Mechanism}}
To verify the effectiveness of the proposed AFM, we conduct the ablation experiment by removing AFM (``w/o AFM"), as shown in the Table 
 \ref{tab:ablationAFM}. 
In addition, we evaluate the importance of adaptive generation of filter strength in AFM by fixing the filter strength $\lambda_{a}$ as 1 and 2, respectively.
We can observe that model with fixed $\lambda_{a}$ performs worse than the model with an adaptive generation mechanism.
This demonstrates that the network can benefit from the adaptive generation of texture filtering strength according to the style of the input image.

\begin{table}[t]
\centering
   \resizebox{\linewidth}{!}{
    \begin{tabular}{c|cccc}
    \toprule
    Settings & Mapillary & BDD-100k & Cityscapes &Avg.\\
    \midrule
    \midrule
    \textbf{w/o} HGGN &39.13  &37.94  &38.91 &38.66\\
    \midrule
    \textbf{w/} 1-level SGE &41.04  &38.61  &40.73 &40.13\\
    \midrule
    \textbf{w/} 2-level SGE &42.72  &40.10  &42.33 &41.72\\
    \midrule
    \textbf{w/} 3-level SGE &43.96 &41.03 &43.25 &42.75\\
    \midrule
    \textbf{w/} 4-level SGE &45.59  & 42.31 &44.83 &44.24\\
    \bottomrule
    \end{tabular}
   }
    \caption{Ablation study of HGGN with SGE modules. The model is trained on the GTA5 with ResNet-101 and validated on three real-world datasets.}
    \vspace{-5pt}
    \label{tab:ablation}
\end{table}

\subsubsection{\textbf{Effect of Hierarchical Guidance Generalization Network}}
\begin{figure}[t]
\centering
\includegraphics[width=1.0\columnwidth]{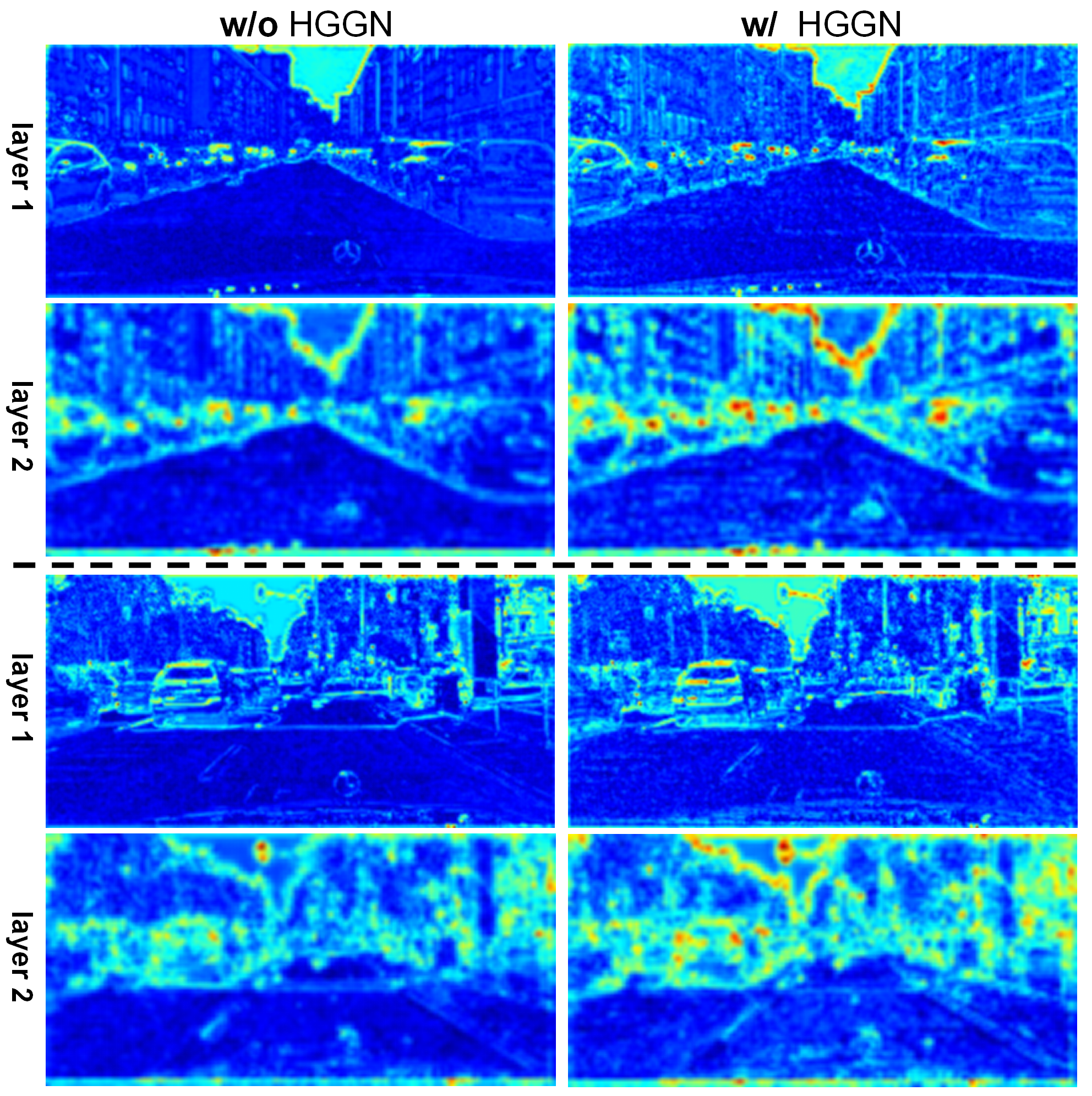} 
\caption{Visual comparison of ``w/o HGGN" and ``w/ HGGN". The right column shows visualized features corresponding to the first and second output features of SGE, while the left column is without SGE.}
\label{visual}
\vspace{-5pt}
\end{figure}

HGGN is designed to extract the domain-invariant feature representations. 
To better understand the extracted domain-invariant feature, we visualize the features under the ``w/ HGGN" and ``w/o HGGN".
As shown in Fig.~\ref{visual}, it is obvious that the features of ``w/ HGGN" focus more on the structure features.
In Table~\ref{tab:ablation}, we verify the effectiveness of the HGGN by conducting the setting of ``w/o HGGN". 
Besides, we deploy different numbers of SGE in the network.
This means that in the absence of SGE, we adopt the corresponding feature layer from the backbone network.
Compared with 4-level SGE, the performance of 1-level SGE (without SGE-2/3/4), 2-level SGE (without SGE-3/4), and 3-level SGE (without SGE-4) on segmentation results are decreased, which indicates that the spatial guidance design after each feature layer can greatly promote generalization.
According to the quantitative evaluation in Table~\ref{tab:ablation} and qualitative results in Fig.~\ref{result}, the HGGN with SGE modules can obtain better generalization performance.

\subsubsection{\textbf{Sensitivity to Hyper-parameter}}
As described above, there is a hyper-parameter $\alpha$ to trade off the importance of the segmentation loss and contour loss in the objective function.
To evaluate the impact of $\alpha$, we set different values to indicate the influence of hyper-parameter on segmentation performance, as shown in Table \ref{tab:ablation1}. The experiments show that the network can obtain better segmentation results when $\alpha$ is set to 2.5.

\begin{table} 
\centering
\resizebox{\linewidth}{!}{
    \begin{tabular}{c|cccc}
    \toprule
    Model (GTA5) & Mapillary & BDD-100k & Cityscapes & Avg. \\
    \midrule
    \midrule
    $\alpha$=1.5 &41.89 &40.72 &42.38 &41.66  \\
    \midrule
    $\alpha$=2.0&43.77 &41.66 &43.61 &43.01 \\
    \midrule
    $\alpha$=2.5 &\textbf{45.59} &\textbf{42.31} &\textbf{44.83}&\textbf{44.24}
 \\
    \midrule
    $\alpha$=3.0 &43.94 &41.75 &43.70 &43.13 \\
    \bottomrule
    \end{tabular}
    }
    \caption{Comparison of mIoU ($\%$) with different $\alpha$ values. The model is trained on the GTA5 with ResNet-101 and validated on three unseen datasets.}
    \vspace{-5pt}
    \label{tab:ablation1}
\end{table}

\section{Conclusion}
This paper studied how to eliminate domain-specific feature influence and seek domain-invariant feature representation in single-domain generalization. 
To alleviate the reliance on domain-specific texture features, we proposed an adaptive filtering mechanism to achieve texture suppression and boost the generalization ability. 
Moreover, a hierarchical guidance generalization network has been designed to guide domain-invariant feature learning, such as spatial and shape information.
Extensive experiments have been presented to reveal the effectiveness of our method.
Our method shows its superiority over other state-of-the-art on the semantic segmentation task. 

\section{Acknowledgments}
This work was supported by the National Natural Science Foundation of China under Grant no. 62072327.

\bibliography{aaai23}

\end{document}